\documentclass[10pt,twocolumn,letterpaper]{article}

\usepackage{cvpr}
\usepackage{times}
\usepackage{epsfig}
\usepackage{graphicx}
\usepackage{amsmath}
\usepackage{amssymb}

\makeatletter
\DeclareRobustCommand*\textsubscript[1]{%
  \@textsubscript{\selectfont#1}}
\def\@textsubscript#1{%
  {\m@th\ensuremath{_{\mbox{\fontsize\sf@size\z@#1}}}}}
\makeatother

\cvprfinalcopy 

\def\cvprPaperID{2789} 

\ifcvprfinal\fi
\begin{document}

\title{Deep Multitask Architecture for Integrated 2D and 3D Human Sensing}

\newcommand*\samethanks[1][\value{footnote}]{\footnotemark[#1]}
\author{Alin-Ionut Popa$^2$\thanks{Authors contributed equally},    Mihai Zanfir$^2$\samethanks,   Cristian Sminchisescu$^{1, 2}$\\
{\tt\small alin.popa@imar.ro, mihai.zanfir@imar.ro cristian.sminchisescu@math.lth.se} \\ 
$^1$Department of Mathematics, Faculty of Engineering, Lund University\\ $^2$Institute of Mathematics of the Romanian Academy}

\maketitle

\begin{abstract}

We propose a deep multitask architecture for \emph{fully automatic 2d and 3d human sensing} (DMHS), including \emph{recognition and reconstruction}, in \emph{monocular images}. The system computes the figure-ground segmentation, semantically identifies the human body parts at pixel level, and estimates the 2d and 3d pose of the person. The model supports the joint training of all components by means of multi-task losses where early processing stages recursively feed into advanced ones for increasingly complex calculations, accuracy and robustness. The design allows us to tie a complete training protocol, by taking advantage of multiple datasets that would otherwise restrictively cover only some of the model components: complex 2d image data with no body part labeling and without associated 3d ground truth, or complex 3d data with limited 2d background variability. In detailed experiments based on several challenging 2d and 3d datasets (LSP, HumanEva, Human3.6M), we evaluate the sub-structures of the model, the effect of various types of training data in the multitask loss, and demonstrate that state-of-the-art results can be achieved at all processing levels. We also show that in the wild our monocular RGB architecture is perceptually competitive to a state-of-the art (commercial) Kinect system based on RGB-D data.

\end{abstract}

\section{Introduction}

The visual analysis of humans has a broad spectrum of applications as diverse as autonomous vehicles, robotics, human-computer interaction, virtual reality, and digital libraries, among others. The problem is challenging due to the large variety of human poses and body proportions, occlusion, and the diversity of scenes, angles of observation, and backgrounds humans are pictured against. The \emph{monocular} case, which is central and intrinsic in many scenarios like the analysis of photographs or video available on the web, adds complexity as depth information is missing for 3d reconstruction. This leads to geometric ambiguity and occlusion which are difficult to palliate compared to situations where multiple cameras are present. 

A detailed analysis at both 2d and 3d levels, further exposes the need for both measurement and prior-knowledge, and the necessary inter-play between segmentation, reconstruction, and recognition within models that can jointly perform all tasks. 

As training is essential, a major difficulty is the limited coverage of current datasets: 2d repositories like LSP \cite{Johnson10} or MPI-II \cite{andriluka14cvpr} exhibit challenging backgrounds, human body proportions, clothing, and poses, but offer single views, carry only approximate 2d joint location ground truth, and do not carry human segmentation or body part labeling information. Their size is also relatively small by today's deep learning standards. In contrast, 3d datasets like HumanEva \cite{sigal2010} or Human3.6M \cite{Ionescu14pami} offer extremely accurate 2d and 3d anatomical joint or body surface reconstructions and a variety of poses captured under multiple viewpoints. Human 3.6M is also large-scale. However, being captured indoors, the 3d datasets typically lack the background and clothing variability that represent a strength of the 2d datasets captured in the wild.\footnote{The situation is slightly more nuanced as some of the 3d datasets (e.g. Human3.6M) come with mixed-reality training setups where a moderately realistic graphics character was placed, in a geometrically correct setup, into a real scene and animated using human motion capture -- arguably, though, a complete, fully realistic 2d and 3d training setting is still elusive.}
An open question is how one can leverage the separate strengths of existing 2d and 3d datasets towards training a model that can operate in realistic images and can offer accurate recognition and reconstruction estimates. 

In this paper we propose one such deep learning model which, given a monocular RGB image, is able to fully automatically sense the humans at multiple levels of detail: figure-ground segmentation, body-part labeling at pixel level, as well as 2d and 3d pose estimation. By designing multi-task loss functions at different, recursive processing stages (human body joint detection and 2d pose estimation, semantic body part segmentation, 3d reconstruction) we are able to tie complete, realistic training scenarios by taking advantage of multiple datasets that would otherwise restrictively cover only some of the model component training (complex 2d image data with no body part labeling and without associated 3d ground truth, or complex 3d data with limited 2d background variability), leading to covariate shift and a lack of model expressiveness. In extensive experiments, including ablation studies performed using representative 2d and 3d datasets like LSP, HumanEva, or Human3.6M, we illustrate the model and show that state-of-the-art results can be achieved for both semantic body part segmentation and for 3d pose estimation.

\section{Related Work}

This work relates to 2d and 3d monocular human pose estimation methods as well as semantic segmentation using fully-trainable deep processing architectures. As prior-work is comprehensive in each sub-domain \cite{rosenhahn08}, we will here mostly cover some of the key techniques directly relating to our approach, with an emphasis towards deep architectures and methodologies aiming to integrate the different levels of 2d and 3d processing.

The problem of 2d human pose estimation has been approached initially using the pictorial structures and deformable part models where the  kinematic tree structure of the human body offers a natural decomposition \cite{felzenszwalb2010object,johnson2010,bourdev10,yang2013,pishchulin2013,HernandezVelaS16}. In recent years, the deep learning methodology had a great impact over the state-of-the-art pose estimation models where different hierarchical feature extraction architectures have been combined with spatial constraints between the human body parts \cite{toshev2014,chen2014,tompson2014joint,kiefel2014human,pons2014,pfister2015flowing, liang2015human,bulat2016,insafutdinov2016,carreira2015human,tang2016multi,newell2016,Lifshitz2016}. Recent deep architectures are obtained by cascading processing stages with similar structure but different parameters, where the system combines outputs of early layers with new information extracted directly from the image using learnt feature extractors. Such recursive schemes for 2d pose estimation appear in the work of \cite{ramakrishna2014,wei2016,carreira2015human} whereas a similar idea of feeding-back 3d estimates into 2d inference layers appears in \cite{Ionescu14}. Note that the iterative processing frameworks of \cite{ramakrishna2014,Ionescu14} are not deep, built on parts-based graphical models and random forests, respectively.

There is a vast literature on monocular 3d human pose estimation, including the analysis of kinematic ambiguities associated with the 3d pose \cite{sminchisescu_cvpr03,bo2008fast}, as well as generative and discriminative methods for learning and inference \cite{rosales2001learning,ShakhnarovichVD03,bo2010twin,agarwal2006recovering,mori2006recovering,sigal07,sminchisescu_locnpred09,pons2014}. 
More recently, deep convolutional architectures have been employed in order to estimate 3d pose directly from images\cite{li20143d,li2015maximum,rhodin2016general,zhou2015sparseness} mostly in connection with 3d human motion capture datasets like HumanEva, Human3.6M, where the poses are challenging but the backgrounds are relatively simple. There is also interest in combining 2d and 3d estimation methods in order to obtain models capable of multiple task, and able to operate in realistic imaging conditions  \cite{ramakrishna2012reconstructing,Wang_2014_CVPR,akhter2015pose}. The most recent methods, \cite{bogo2016keep,rhodin2016general}, rely on an a-priori 3d model that is fitted to anatomical body joint data obtained from an initial 2d pose estimate produced by a deep processing architecture, like the one of \cite{insafutdinov2016} or \cite{tompson2014joint}. The methods rely on a state-of-the-art discriminative person detection and 2d pose estimation and introduce a generative fitting component to search in a space of admissible body proportion variations. Both approaches fit a statistical body shape and kinematic model to data, using one or multiple views and the joint assignment constraints from 2d human pose estimation. These methods use the 3d to 2d anatomical landmark assignment provided by \cite{bogo2016keep,rhodin2016general} as initialization for 3d human pose estimation. While this is effective, as shown in several challenging evaluation scenarios, the initial 3d shape and kinematic configuration still needs to be initialized manually or set to a key initial pose. This is in principle still prone to local optima as the monocular 3d human pose estimation cost is non-linear and non-convex even under perfect 3d to 2d model-image assignments \cite{sminchisescu_cvpr03}. 

We share with \cite{Ionescu14,bogo2016keep,rhodin2016general} the interest in building models that integrate 2d and 3d reasoning. We propose a fully trainable discriminative model for human recognition and reconstruction at 2d and 3d levels. We do not estimate human body shape, but we do estimate figure-ground segmentation, the semantic segmentation of the human body parts, as well as the 2d and 3d pose. The system is trainable, end-to-end, by means of multitask losses that can leverage the complementary properties of existing 2d and 3d human datasets. The model is fully automatic in the sense that both the human detection and body part segmentation and the 2d and 3d estimates are the result of recurrent stages of processing in a homogeneous, easy to understand and computationally efficient architecture. Our approach is complementary to \cite{bogo2016keep,rhodin2016general}: our model can benefit from a final optimization-based refinement and it would be useful to estimate the human body shape. In contrast, \cite{bogo2016keep,rhodin2016general} can benefit from the semantic segmentation of the human body parts for their shape fitting, and could use the accurate fully automatic 2d and 3d pose estimates we produce as initialization for their 3d to 2d refinement process.

\section{Methodology}

In this section we present our multitask multistage architecture. The idea of using multiple stages of recurrent feed-forward processing is inspired by architectures like \cite{ramakrishna2014,wei2016,carreira2015human} which focus separately on the 2d and 3d domains. However, we propose an uniform architecture for joint 2d and 3d processing that no prior method covers. Our choice of multi-task loss also makes it possible to exploit the complementary advantages of different datasets. 

\begin{figure}[!htbp]
    \begin{center}
        \includegraphics[width=0.7\linewidth]{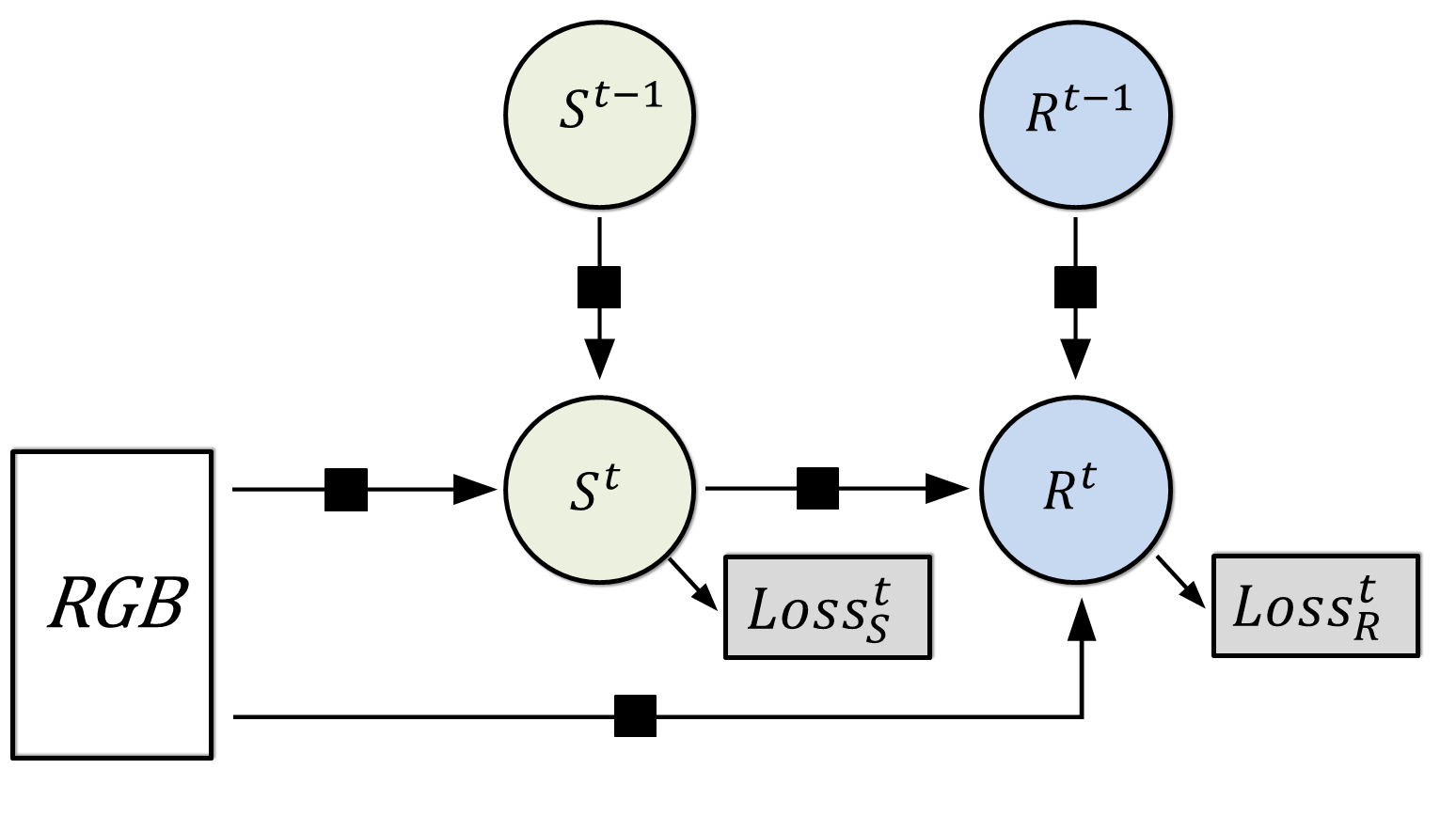}
    \end{center}
\caption{Stage $t$ of our architecture for recognition and reconstruction: figure-ground segmentation, 2d pose estimation, and semantic segmentation of body parts, all denoted by $(S)$ and 3d reconstruction $(R)$. The semantic task is detailed in fig. \ref{fig:architectureFigure2} and fig. \ref{fig:labelingModule}; the 3d reconstruction task is detailed in fig. \ref{fig:reconstructionModule}.}
\label{fig:architectureFigure1}
\end{figure}

\begin{figure}[!htbp]
    \begin{center}         
            \includegraphics[width=0.7\linewidth]{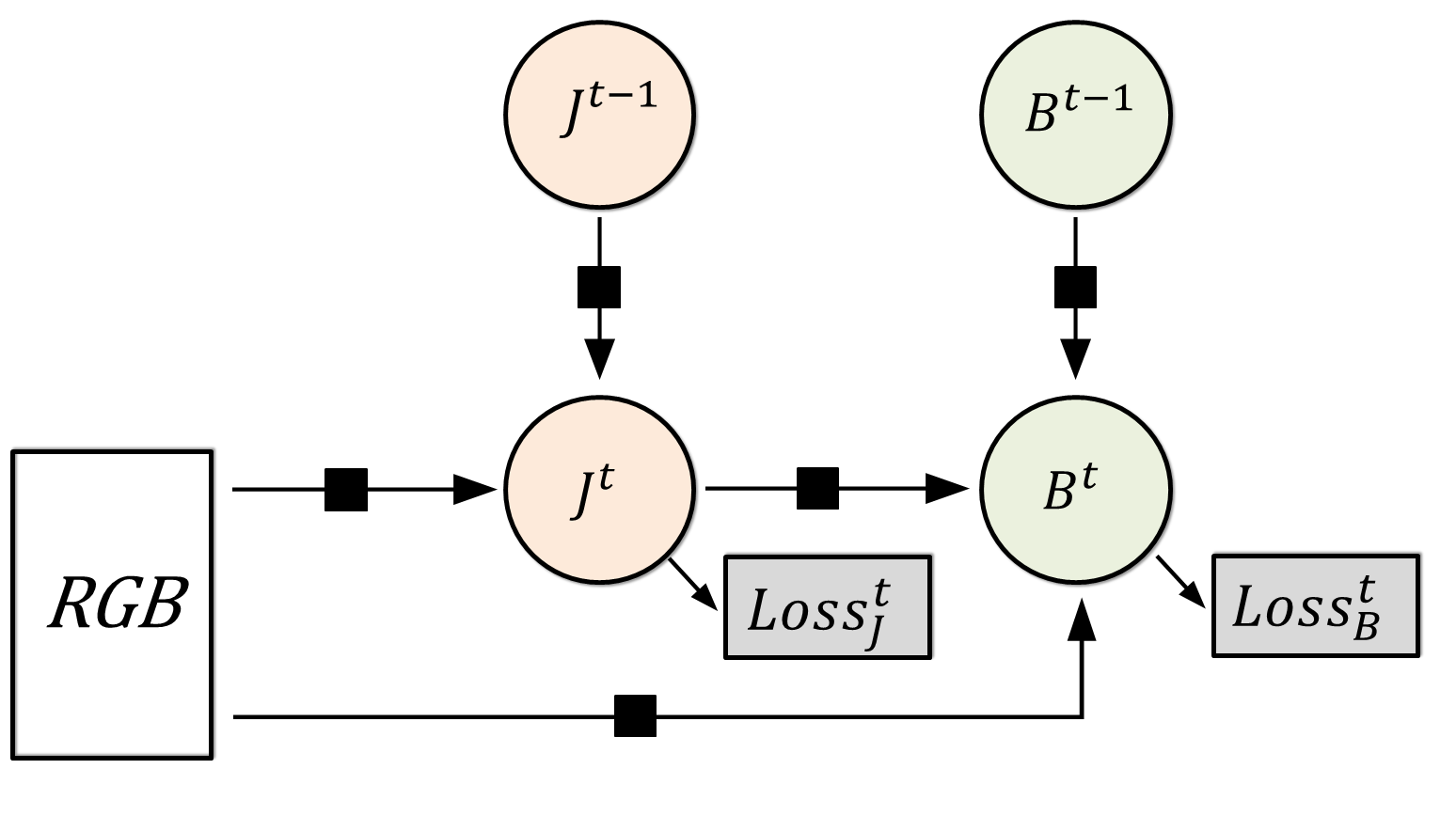}
      
    \end{center}
\caption{Stage $t$ of our semantic task, including 2d joint detection $(J)$, and labeling of the body parts $(B)$.}
\label{fig:architectureFigure2}
\end{figure}

Conceptually, each stage of processing in our model produces recognition and reconstruction estimates and is constrained by specific training losses. Specifically, each stage $t$ is split into semantic processing $S^{t}$, and 3d reconstruction, $R^t$ (see fig. \ref{fig:architectureFigure1}). At the same time, the semantic module $S^t$ is divided in two sub-tasks, one focusing on 2d pose estimation, $J^t$, and the other on body part labeling and figure-ground segmentation, $B^t$ (see fig. \ref{fig:architectureFigure2}). The first one (\textit{i.e.} $J^t$) feeds into the second one (\textit{i.e.} $B^t$), while the semantic stages feed into the reconstruction stages. Each task consists of a total of six recurrent stages which take as input the image, the results of the previous stages of the same type (except for the first one), as well as inputs from other stages (2d pose estimation feeding into semantic body part segmentation and both feeding into 3d pose reconstruction). The inputs to each stage are individually processed and fused via convolutional networks in order to produce the corresponding outputs. 

\subsection{2D Human Body Joint Detection}\label{sec:2dpose} 

The 2d pose estimation task is based on a recurrent convolutional architecture similar to \cite{wei2016}. Given an RGB image $I \in \mathbb{R}^{w\times h\times 3}$, we seek to correctly predict the locations of $N_J$ anatomically defined human body joints $p_k \in \mathcal{Z} \subset \mathbb{R}^2$, with $k \in{\{1\dots N_J}\}$. At each stage $t \in{\{1\dots T}\}$, where $T$ is the total number of stages, the network outputs belief maps $J^t \in \mathbb{R}^{w \times h \times N_J}$. The first stage of processing operates only on image evidence (a set of seven convolution and three pooling layers producing features ${\bf x}$) but for subsequent stages, the network also considers the information in the belief maps fed from the previous stage $J^{t-1}$ with a slightly different image feature function ${\bf x'}$, defined as a set of four convolution and three pooling layers as in \cite{wei2016}. These features are transformed through a classification function $c^t_J$ to predict the body joint belief maps $J^t$. The function $c^t_J$ consists of a series of five convolutional layers, the first three of the form $(11\times 11\times 128)$, followed by a $(1\times 1\times 128)$ convolution and a final $(1\times 1\times N_J)$ convolution that outputs $J^t$.
The loss function at each stage $L^t_{J}$ minimizes the squared Euclidean distance between the predicted and ground truth belief maps, $J^t$ and $J^\star$ :

\begin{equation}
    L^t_{J} = \sum^{N_J}_{k=1}\sum^{ }_{z \in\mathcal{Z}}\left \| J^t(z,k) - J^\star(z,k)\right \|_2^2
\end{equation}

In practice, this component of the model can be trained with data from both 2d and 3d datasets like LSP, where the ground truth is obtained manually, or HumanEva and Human3.6M where the ground truth is obtained automatically based on anatomical markers.

\subsection{Semantic Body Part Segmentation} 

In semantic body part segmentation (body part labeling) we assign each image location $(u,v) \in \mathcal{Z} \subset \mathbb{R}^2$ one of $N_B$ anatomical body part labels (including an additional label for background), $b_l$, where $l \in{\{1\dots N_B}\}$. At each stage $t$, the network predicts, for each pixel location, the probability of each body part being present, $B^t \in \mathbb{R}^{w\times h\times N_B}$. Differently from the previous task, our aim now is to classify each pixel location, not only to identify the body joints. The loss function used changes from a squared Euclidean loss to a multinomial logistic loss:
\begin{equation}
    L^t_{B} = -\frac{1}{|\mathcal{Z}|}\sum^{ }_{z \in\mathcal{Z}}\log (B^t_{z,B^{\star}_z}) 
\end{equation}
where $B^{\star}_z$ is the ground truth label for each image location $z=(u,v)$.

\begin{figure*}
    \begin{center}
            \includegraphics[width=0.85\linewidth, height=6.7cm]{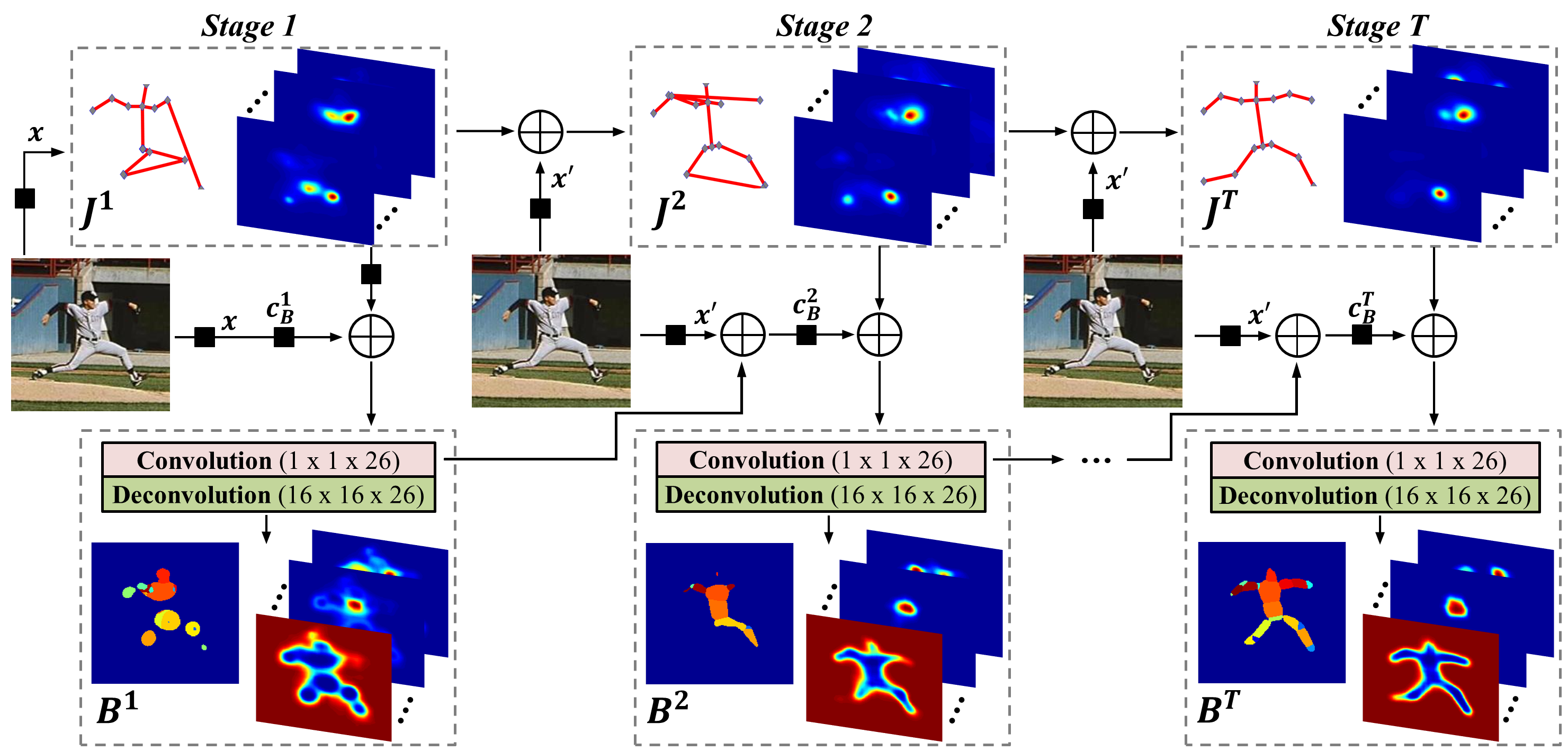} 
    \end{center}
\caption{Our multitask multistage 2d semantic module $S^t$, combines semantic body part labeling $B^t$ and 2d pose estimation $J^t$.}
\label{fig:labelingModule}
\end{figure*}

 During the first stage of processing, we use convolutional representations based on the image (a series of convolution and pooling layers ${\bf x}$ with parameters tied from \S\ref{sec:2dpose}) and the 2d pose belief maps $J^1$ in order to predict the current body labels $B^1$. For each of the following stages, we also use the information present in the body labels at the previous stage, $B^{t-1}$, and rely on a series of four convolutional layers $c^t_B$ that learn to combine inputs obtained by stacking image features ${\bf x}$ and $B^{t-1}$. The function $c^t_B$ shares the same structure as the first four convolutions in $c^t_J$, but a classifier in the form of a $(1\times 1\times N_B)$ convolution is applied after the fusion with the current 2d pose belief maps $J^t$, in order to obtain semantic probability maps $B^t$. An overview of our architecture together with the main dependencies is given in figure \ref{fig:labelingModule}. Finally, we use an additional deconvolution layer \cite{Long15} of size $16\times16\times N_B$, such that the loss can be computed at the full resolution of the input image $I$. 
 
 In practice, realistic data for training this component of the loss is not as easy to obtain as 2d body joint positions. Human3.6M offers such training data, but we are also able to generate it approximately for LSP (see \S\ref{sec:exps}).

\subsection{3D Pose Reconstruction} 

\begin{figure*}
    \begin{center}
            \includegraphics[width=0.85\linewidth, height=9.3cm]{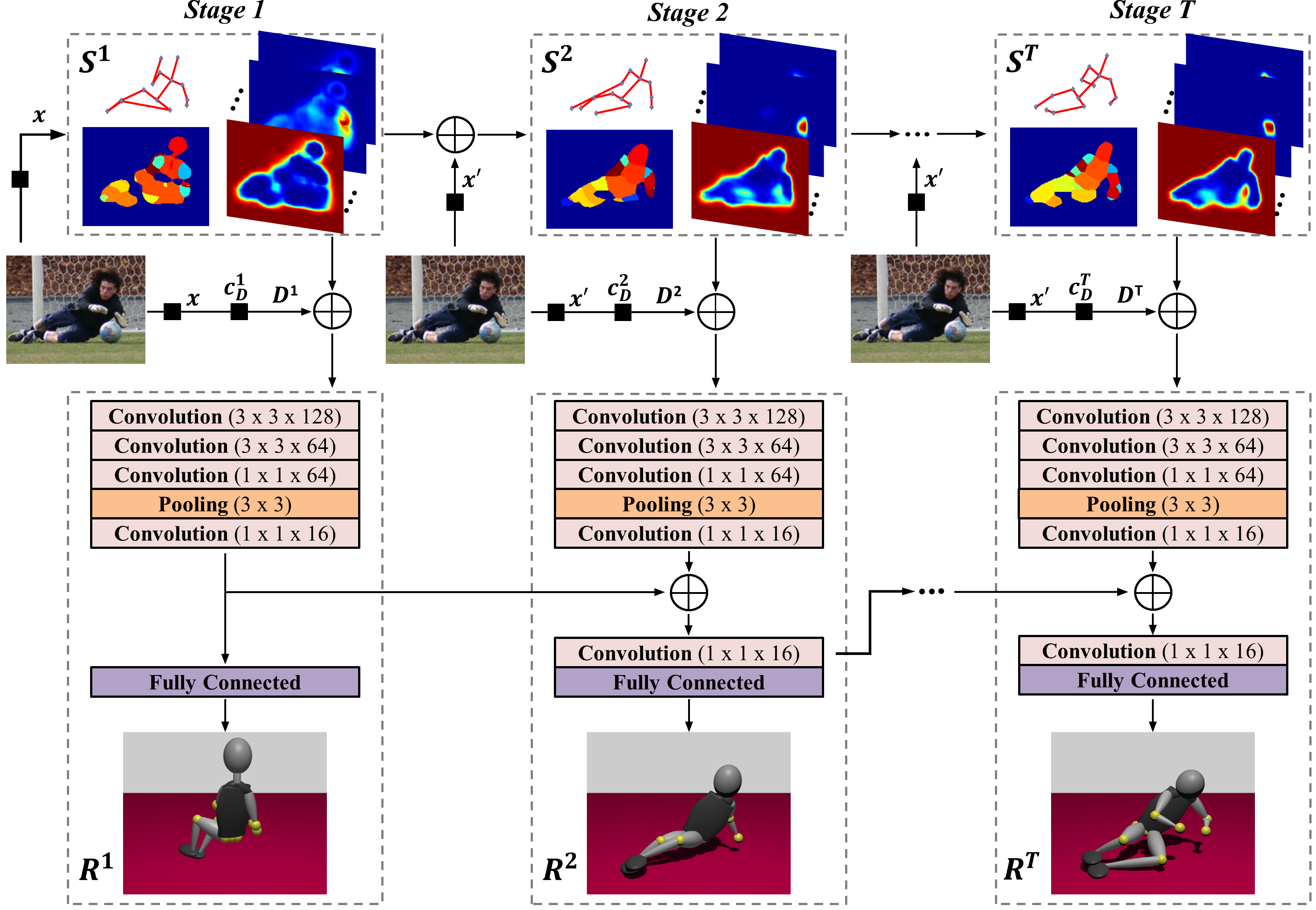} 
    \end{center}
\caption{Our multitask multistage 3d reconstruction module $R^t$, combines 3d processing with information from semantic modules, $S^t$.}
\label{fig:reconstructionModule}
\end{figure*}

This model component is designed for the stage-wise, recurrent reconstruction of 3d human body configurations represented as set of $N_R$ 3d skeleton joints, from a single monocular image $I$. The estimate is obtained from the internal representations $R^t$. The 3d reconstruction module leverages information provided by the 2d semantic components $S^t$, incorporating the joint and body part labeling feature maps $J^t$ and $B^t$. Additionally, we insert a trainable function $c^t_D$, defined similarly to $c^t_B$, over image features, in order to obtain body reconstruction feature maps $D^t$. The module follows a similar flow as the previous ones: it reuses estimates at earlier processing stages, $R^{t-1}$, together with $S^t$ and $D^t$, in order to predict the reconstruction feature maps $R^t$. The processing stages and dependencies of this module are shown in fig. \ref{fig:reconstructionModule}.

Procedurally, we first fuse $S^t$ and $D^t$, then apply a series of one $(3\times 3\times 128)$ convolution, one $(3\times 3\times 64)$ convolution,  one $(1\times 1\times 64)$ convolution, followed by a pooling layer $(3\times 3)$ and a $(1\times 1\times 16)$ convolution. The output is concatenated with $R^{t-1}$ and convolved by a $(1\times 1\times 16)$ kernel that learns to combine the two components, producing the estimate $R^t$. The feature maps are then transformed to the desired dimensionality of the 3d human body skeleton by means of a fully connected layer. The loss $L^t_R$ is expressed as the mean per joint
position error (MPJPE):

\begin{equation}
    L^t_{R} = \sum^{N_R}_{i = 1}\sqrt{\sum^{3}_{j=1}(f(R^t,i,j)-R^\star(i,j))^2 + \epsilon^2} 
\end{equation}
where $R^{\star}$ are the 3d ground truth human joint positions, $f(\cdot)$ is the fully connected layer applied to $R^t$ and $\epsilon$ is a small constant that makes the loss differentiable.

This loss component can be trained with data from HumanEva and Human3.6M, but not from LSP or other 2d datasets as these lack 3d ground truth information. Although the backgrounds in HumanEva and Human3.6M are not as challenging as those in LSP, the use of a multitask loss makes the complete 2d and 3d model competitive not only in a laboratory setting but also in the wild (see \S\ref{sec:exps} and fig. \ref{fig:sampleResults}).

\subsection{Integrated Multi-task Multi-stage Loss}

Given the information provided in the previous sections, we are now able to define the complete multitask, multistage loss function of the model as follows:

\begin{equation}\label{eq:mtloss}
L=\sum_{t=1}^T (L^t_{J}+L^t_{B}+L^t_{R})
\end{equation}

The loss allows us to conveniently train all the model component parameters, for different tasks, based on datasets where the annotations are challenging, or where annotations are missing entirely, as datasets with different coverage contribute to various components of the loss. Whenever we train using datasets with partial coverage, we could freeze the model components for which we do not have ground-truth. We can also simultaneously train all parameters, using datasets with partial and complete coverage: those examples for which we have ground truth at all levels will contribute to each of the loss components, whereas examples for which we have partial ground truth will only contribute to their corresponding losses.

\section{Experiments}\label{sec:exps}

In order to evaluate our method, we use $3$ well-known datasets, the Leeds Sports Dataset (LSP) \cite{Johnson10}, HumanEva \cite{sigal2010} and Human3.6M \cite{Ionescu14pami}. 

The LSP dataset consists of 2d pose annotated RGB images depicting sports people (\textit{athletics, badminton, baseball, gymnastics, parkour, soccer, tennis, volleyball}). We employ both the original release containing $1,000$ training images and $1,000$ testing images, as well as the extended training release containing an additional $10,000$ images.  

We use the first release of the HumanEva(-I) dataset. This dataset was captured by an accurate 3d motion capture system in a laboratory environment. There are six actions performed in total by three different subjects. As standard procedure \cite{simo2013joint,bo2010twin,Yasin2016,Tekin2016}, we train our model on the train set and report results on the validation set, where we only consider every $5^{th}$ frame of the sequences \textit{walking}, \textit{jog} and \textit{box} for all three subjects and a single frontal camera view.

Human80K is an 80,000 sample subset of the much larger 3.6 million human pose dataset Human3.6M \cite{Ionescu14pami}. The dataset is captured in a laboratory environment with a motion capture setup, and contains daily activities and interaction scenarios (\textit{providing directions, discussion, eating, activities while seating, greeting, taking photo, posing, making purchases, smoking, waiting, walking, sitting on chair, talking on the phone, walking dog, walking together}). The actions are performed by $11$ actors, and captured by $4$ RGB cameras. The dataset is fully annotated and it contains RGB data, 2d body part labeling ground truth masks as well as accurate 2d and 3d pose reconstructions. Human80K consists of $55,144$ training and $24,416$ testing samples from Human3.6M. The samples from each original capture were selected such that the distance between each-other, in 3d space, is no more than 100 mm. 

We use the Caffe \cite{jia2014caffe} framework to train and test our architecture. The complete recognition and reconstruction pipeline takes approximately 400 ms per frame, in testing, on an Nvidia TITAN X (Pascal) 12GB GPU. We evaluate the recognition (2d body part labeling) and 3d reconstruction capabilities of our architecture. We use $T = 6$ stages in our architecture for each sub-task component model (joint detection, semantic segmentation, 3d reconstruction) and report results only for the final stage of each sub-task, as it is the best performing according to validation.

\subsection{Body Part Labeling}

In order to evaluate the 2d body part labeling task, we use the Human80K and LSP datasets. We introduce additional annotations for LSP, as they are not available with the original release of the dataset which only provides the 2d anatomical joints. We create human body part annotations for LSP by using the annotated 2d skeleton joints and the kinematic tree. We  produce circles for skeleton joints and ellipses for the individual body parts. We set the major axis to the size of the segment between the corresponding joints and estimate a constant value, for each body part, for the minor axis. The resulting masks (filtered by visual inspection) are of lower quality than those available for Human3.6M. The reason for augmenting LSP, is to enhance the variability in human appearance, body proportions and backgrounds in Human80K. Thus, we want to leverage the good quality labeling and pose variation of Human80K with the diverse appearance variations found in LSP.

For evaluation on Human80K, we compare with the results of \cite{Ionescu14} which represent the state-of-the-art for this task on this dataset. The authors of \cite{Ionescu14} assume that the silhouette of the person (the figure-ground segmentation) is given, and perform the body part labeling only on the foreground mask as an inference problem over a total of $24$ unique labels. Different from them, we do not make this assumption and consider the background as an extra class, thus building a model that predicts $25$ classes.

To extend the evaluation on LSP, we consider multiple scenarios: (a) training on Human80K with (b) fine-tuning on LSP and (c) training our architecture on LSP and Human80K simultaneously and test on both Human80K and LSP. In our setup, training using only LSP was not feasible, as in multiple experiments our network failed to converge. The labeling parameters (stages $B$) are initialized randomly, while the parameters corresponding to the 2d joint detection components, $J$, are initialized with the values of the network presented in \cite{wei2016}, trained on MPI-II and LSP.

The performance of our body part labeling models for the Human80K and LSP is given in tables \ref{tbl:performanceSemanticH80K} and \ref{tbl:performanceSemanticLSP}, respectively. We use the same evaluation metrics as in \cite{Ionescu14}, \textit{i.e.} average accuracy for the pixels contained within the ground truth silhouette and class normalized average accuracy. Basically, these two metrics apply only to the foreground classes. Additionally, we compute for all pixels, background and foreground, the average accuracy as well as the class normalized average accuracy.

\begin{table*}[!htbp]
\begin{center}
\scalebox{0.88}{\begin{tabular}{|l||c|c|c|c|}
\hline
Metric 		& DMHS\textsubscript{B} - Human80K	& DMHS\textsubscript{B} - LSP (ft) & DMHS\textsubscript{B} - Human80K \& LSP & \cite{Ionescu14}	\\
\hline
\hline
Avg. Acc. (\%) per pixel (fg) &	\textbf{79.00}	& 	53.31	& 75.84	 &	73.99\\
\hline 
Avg. Acc. (\%) per pixel (fg + bg) &	\textbf{91.15}	& 	83.30 & 	89.92 & -	\\
\hline 
Avg. Acc. (\%) per class (fg) 	& \textbf{67.35}	& 	43.40	&  64.83	 & -	\\
\hline 
Avg. Acc. (\%) per class (fg + bg) 	& \textbf{68.56}	& 	45.61	&  66.13	 & 53.10	\\
\hline
\end{tabular}}
\end{center}
\caption{\small{\textbf{Body part labeling results for the Human80K test set}. We report the performance of our network, trained on Human80K, LSP and both datasets jointly. Note that for LSP, the network was pre-trained on Human80K, as it otherwise failed to converge trained on LSP alone. We also compare with \cite{Ionescu14}, where comparisons are possible only for accuracies computed on the person foreground as the model of \cite{Ionescu14} does not predict the background label. Our model is able to predict the background class, thus we report the performance for the entire image (including the background class) as well as performance inside the silhouette (for classes associated to human body parts). Note that the best performance on Human80K is obtained with the network trained on Human80K, perhaps due to the noisy annotations added to LSP, and the naturally more similar training and testing distributions of Human80K.}} \label{tbl:performanceSemanticH80K}\vspace{-.2cm}
\end{table*}

From table \ref{tbl:performanceSemanticH80K} (Human80K testing), it can be noted that even though we solve a harder problem (by additionally estimating the figure-ground segmentation), the class normalized average precision greatly improves, with more than 10\% over \cite{Ionescu14} for the models trained on (a) Human80K and (c) Human80K and LSP jointly. However, the model (b) initialized with parameters trained on Human80K, but fine-tuned on LSP, seems to have a performance drop caused by the low quality of the LSP body label annotations. In this scenario, as expected, the best performance is obtained by the model trained on Human80K, perhaps due to the fact that the test set distribution is better captured by the unaltered Human80K training set. This is not the case when testing on LSP, as it can be seen from table \ref{tbl:performanceSemanticLSP}. The best performance is obtained by the model trained jointly on Human80K and LSP. This model is able to combine the pose variability and part labeling annotation quality of Human80K with the background and appearance variability of LSP, making it adequate for both laboratory settings and for images of people captured against challenging backgrounds.

\begin{table*}[!htbp]
\begin{center}
\scalebox{0.88}{\begin{tabular}{|l||c|c|c|c|}
\hline
Metric 		& DMHS\textsubscript{B} - Human80K	& DMHS\textsubscript{B} - LSP (ft) & DMHS\textsubscript{B} - Human80K \& LSP \\
\hline
\hline
Avg. Acc. (\%) per pixel (fg) &	50.52	& 	60.54	& \textbf{61.16} \\
\hline 
Avg. Acc. (\%) per pixel (fg + bg) &	85.58	& 	91.08	& \textbf{91.09}\\
\hline 
Avg. Acc. (\%) per class (fg) 	& 36.46	& 	44.73	&  \textbf{45.91}	\\
\hline 
Avg. Acc. (\%) per class (fg + bg) 	& 38.77	& 	46.88	&  \textbf{48.01}	\\
\hline
\end{tabular}}
\end{center}
\caption{\small \textbf{Body part labeling results for the LSP dataset}. All models are initialized on the Human80K dataset, as networks trained only using LSP failed to converge. In this case the models trained jointly on Human80K and LSP produced the best results. This shows the importance of having accurate body part labeling annotations, obtained in simple imaging scenarios but for complex body poses, in combination with less accurate annotations but with more complex foreground-background appearance variations.} \label{tbl:performanceSemanticLSP}\vspace{-.2cm}
\end{table*}

The proposed models for the 2d body part labeling task are trained using an initial learning rate set to $10^{-10}$ and reduced at every $5$ epochs by a constant factor $\gamma=0.33$. During the learning process, the training data is augmented by randomly rotating with angles between $[-40^{\circ}, +40^{\circ}]$, scaling by a factor in the range $[0.5, 1.2]$ and horizontally flipping, in order to improve the diversity of the training set. Qualitative results of the semantic body part segmentation in challenging images are shown in fig. \ref{fig:sampleResults}.

\subsection{3D Human Pose Reconstruction}

For the evaluation of the 3d pose reconstruction task, we use HumanEva-I and Human3.6M (specifically, Human80K), as they are both challenging datasets, containing complementary poses, and offering accurate 3d pose annotations. In all experiments, we use the same 3d evaluation metrics as other methods we compare against, and according to the standard practice. For the following experiments, note that our 2d semantic module components are trained on data from LSP and Human80K, whereas the 3d component of the module was pre-trained using Human80K. 

We report results on the Human80K test set and investigate the impact of each of the input features $J$, $B$ and $D$ on the overall performance of the 3d reconstruction task (see table \ref{tbl:performance3DPoseH80K}). Notice that models that use the `$D$-pipeline' for the 3d reconstruction part, would correspond to convolutional networks that feed-forward from the image features, but do not leverage the semantic segmentation of the human body parts. 

We observe that our fully integrated system, DMHS\textsubscript{R}(J,B,D), achieves the lowest error of $63.35$ mm. Besides the interest in computing additional detailed semantic human representations, it can be seen that by feeding in the results from other tasks -- the body joint belief maps $J$ and the body labeling probability maps $B$ -- the error is reduced considerably, from $77.56$ mm (for a model based on feed-forward processing and infering additional 2d joint position information) to $63.35$ mm. Notice also our significant gains with respect to the previous state-of-the-art results on Human80K, as reported by \cite{Ionescu14}.

\begin{table*}[!htbp]
\begin{center}
\scalebox{0.88}{\begin{tabular}{|l||c|c|c|c|c|c|c|c|c|c|c|c|c|}
\hline
Method 		& \multicolumn{3}{|c|}{Walking}	& Avg. & \multicolumn{3}{|c|}{Jog} & Avg. & \multicolumn{3}{|c|}{Box} & Avg. & All \\
\hline
\hline
\cite{simo2013joint} &	65.1    &   48.6    & 73.5 & 62.4 & 74.2 & 46.6 & 32.2 & 51.0 & - & - & - & - & -   \\
\hline
\cite{bo2010twin} &	45.4    &   28.3    & 62.3 & 45.33 & 55.1 & 43.2 & 37.4 & 45.2 & 42.5 & 64.0 & 69.3 & 58.6 & 49.7   \\
\hline
\cite{Yasin2016} &	35.8	& 	32.4	& 41.6 & 36.6 & 46.6 & 41.4 & 35.4 & 41.1 & - & - & - & - & -  \\
\hline 
\cite{Tekin2016} &	37.5	& 	25.1	& 49.2 & 37.3 & - & - & - & - & 50.5 & 61.7 & 57.5 & 56.6 & - \\
\hline 
\hline
DMHS\textsubscript{R}(J,B,D) &	27.1 & 18.4 & 39.5 & \textbf{28.3} & 37.6 & 28.9 & 27.6 & \textbf{31.4} & 30.5 & 45.8 & 48.0 & \textbf{41.5}  & \textbf{33.7}  \\
\hline
\end{tabular}}
\end{center}
\caption{\small \textbf{3d mean joint position error on the HumanEva-I dataset}, computed between our predicted joints and the ground truth after an alignment with a rigid transformation. Comparisons with other competing methods show that DMHS achieves state-of-the-art performance.} \label{tbl:performance3DPoseHEVA}\vspace{-.5cm}
\end{table*}

For HumanEva-I, we use the DMHS\textsubscript{R}(J,B,D) architecture trained on Human80K and fine-tuned for a few epochs ($6$) on the HumanEva-I training data. We use a subset of the training set containing $4,637$ samples. The fine-tuning step on HumanEva-I is performed in order to compensate for the differences in marker positioning with respect to Human80K, and in order to account for the different pose distributions w.r.t Human80K. In table \ref{tbl:performance3DPoseHEVA}, we compare our results against several state-of-the-art methods. We follow the standard evaluation procedure in \cite{simo2013joint,bo2010twin,Yasin2016,Tekin2016} and sample data from the validation set for \textit{walking, jogging and boxing} activities. We use a single camera. We obtain considerable performance gains with respect to the previous state-of-the-art methods on HumanEva, even though we only use a small subset of the available training set. 

\begin{figure*}[!htbp]
    \begin{center}
        \begin{tabular}{cc}
            \includegraphics[width=0.34\linewidth,height=3cm]{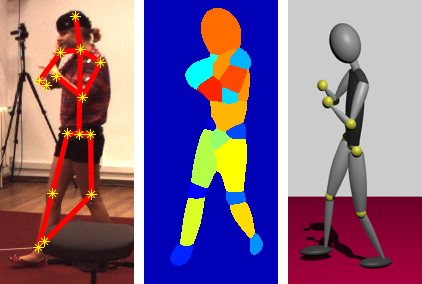} &
            \includegraphics[width=0.34\linewidth,height=3cm]{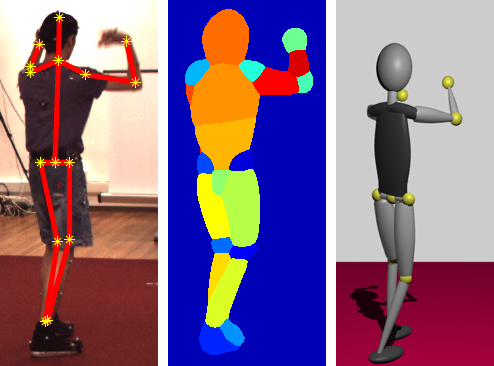} \\
            \includegraphics[width=0.34\linewidth,height=2.14cm]{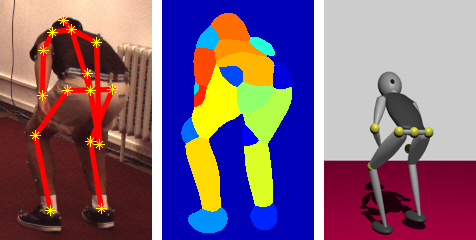} &
            \includegraphics[width=0.34\linewidth,height=2.14cm]{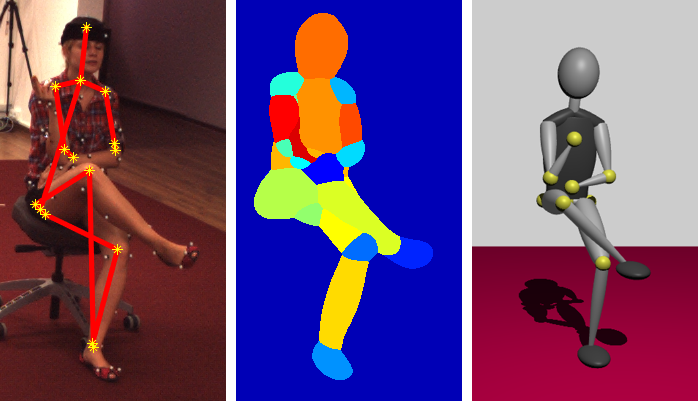}\\
            \includegraphics[width=0.34\linewidth,height=2.14cm]{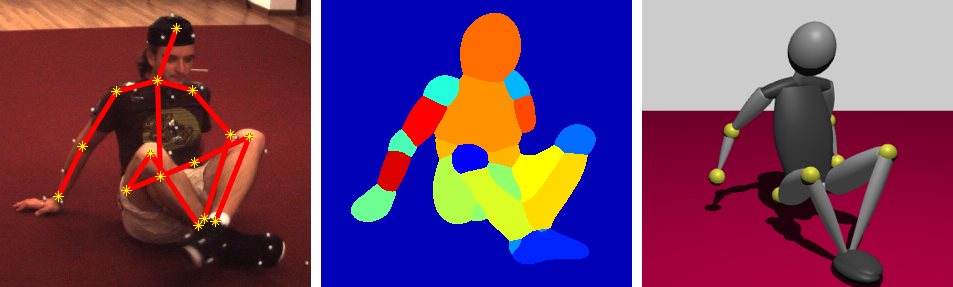} &
            \includegraphics[width=0.34\linewidth,height=2.14cm]{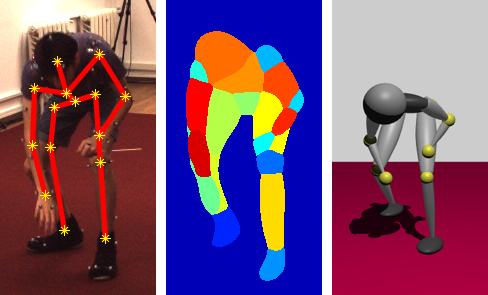} \\

            \includegraphics[width=0.34\linewidth,height=2.14cm]{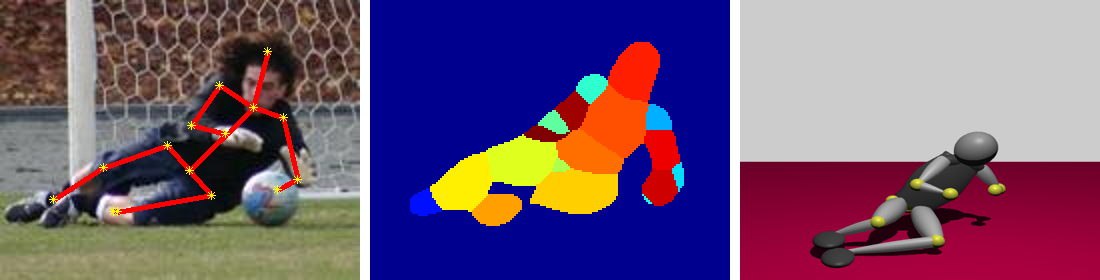} &
            \includegraphics[width=0.34\linewidth,height=2.14cm]{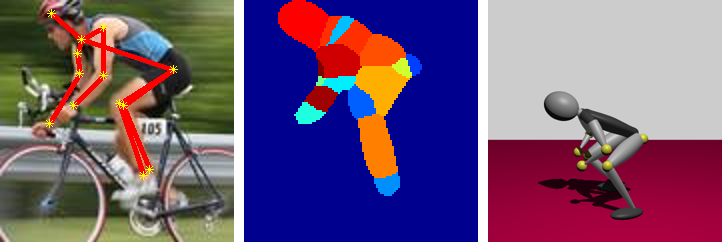} \\
            \includegraphics[width=0.34\linewidth,height=2.14cm]{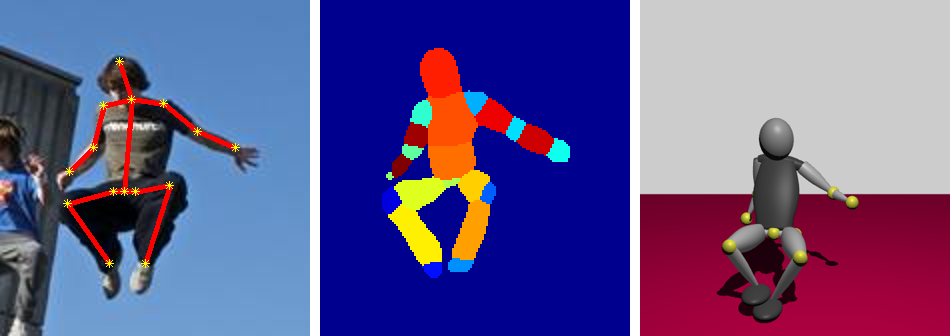} &
            \includegraphics[width=0.34\linewidth,height=2.14cm]{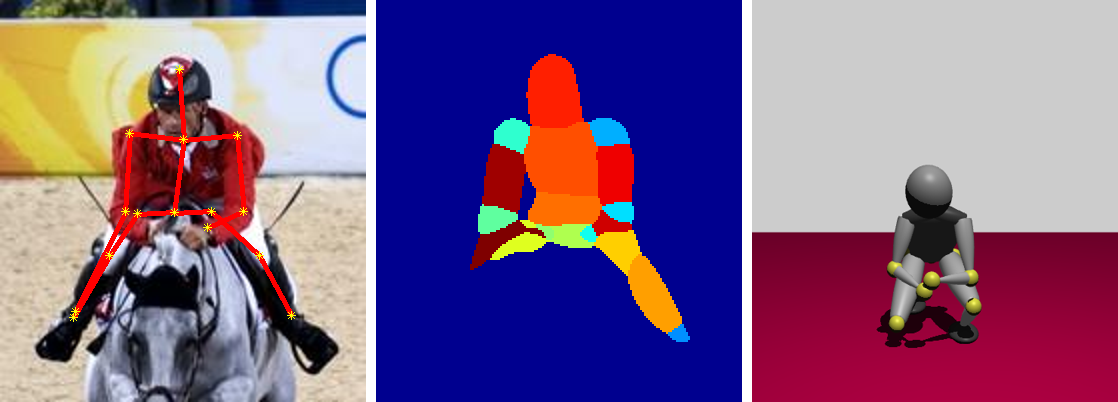}\\
            \includegraphics[width=0.34\linewidth,height=2.14cm]{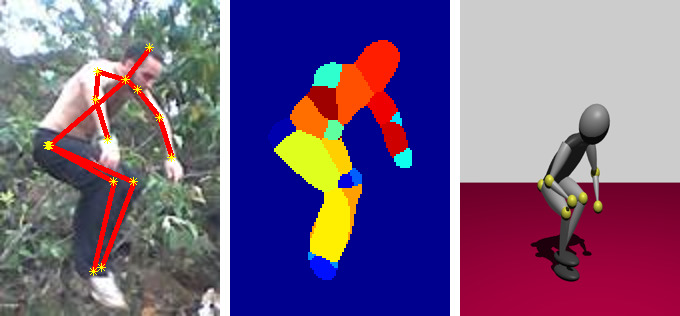} &
            \includegraphics[width=0.34\linewidth,height=2.14cm]{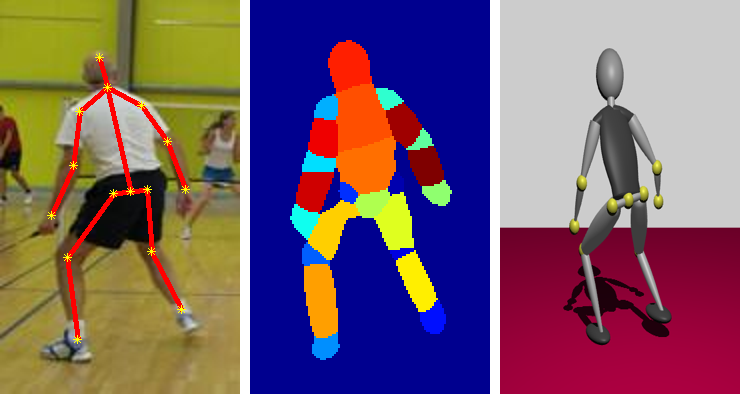}\\
            
        \end{tabular}
        
    \end{center}
\caption{Recognition and reconstruction results for images from Human3.6M and LSP. For each image we show the 2d pose estimate the semantic segmentation of body parts and the 3d pose estimation. Notice the difficulty of backgrounds and poses and the fact that the 2d and 3d models generalize well. Notice that in our architecture, errors during the early stages of processing (2d pose estimation) can be corrected later, during e.g. semantic body part segmentation or 3d pose estimation.}
\label{fig:sampleResults}
\end{figure*}

\begin{figure*}[!htbp]
\begin{center}
        \begin{tabular}{c}
           \includegraphics[width=1\linewidth,height=4.3cm]{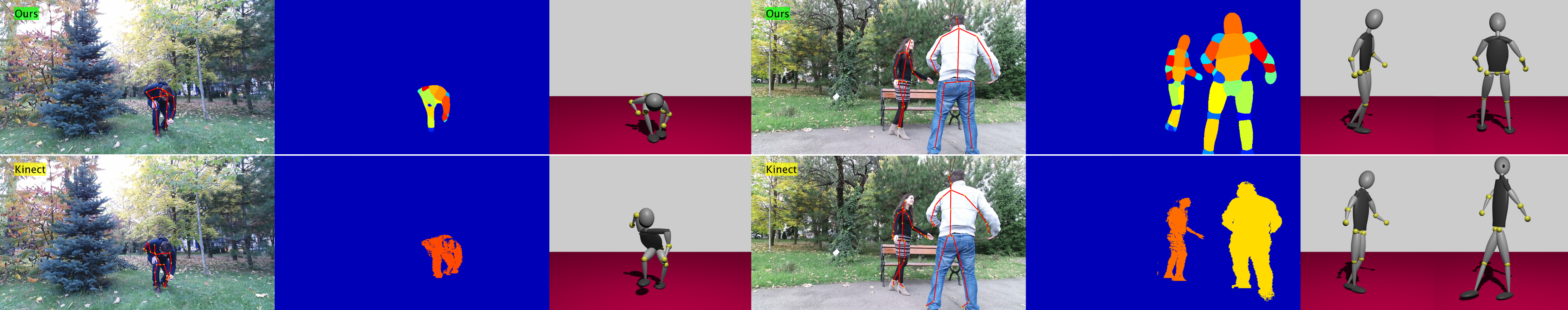}  
        \end{tabular}
    \end{center}
    \caption{Qualitative comparisons for segmentation and reconstruction between our RGB model (top row) and the ones of a commercial RGB-D Kinect for Xbox One system (bottom row). Our model produces accurate figure-ground segmentations, body part labeling, and 3d reconstruction for some challenging poses.}
\label{fig:KinectSampleResults}
\end{figure*}

    For this model, we use an initial learning rate set to $10^{-7}$ and reduce it at every $5$ epochs by a constant factor $\gamma=0.66$. Qualitative results of our method can be seen in fig. \ref{fig:sampleResults} and fig. \ref{fig:KinectSampleResults}.

\begin{table}[!htbp]
\begin{center}
\scalebox{0.88}{\begin{tabular}{|l||c|}
\hline
Model 		& Avg. MPJPE (mm)	\\
\hline
\hline
\cite{Ionescu14}			&	92.00 \\
\hline
DMHS\textsubscript{R}(J)		& 	128.05 \\
\hline
DMHS\textsubscript{R}(D) & 77.56 \\
\hline 
DMHS\textsubscript{R}(J,B)	&  118.68 \\
\hline 
DMHS\textsubscript{R}(J,D)	&  72.00 \\
\hline
DMHS\textsubscript{R}(J,B,D)	&  \textbf{63.35} \\
\hline
\end{tabular}}
\end{center}
\caption{\small \textbf{3d mean joint position error on the Human80K dataset}. Different components of our model are compared to \cite{Ionescu14}.} 
\label{tbl:performance3DPoseH80K}\vspace{-.5cm}
\end{table}

\section{Conclusions}

We have proposed a deep multitask architecture for \emph{fully automatic 2d and 3d human sensing} (DMHS), including \emph{recognition and reconstruction}, based on \emph{monocular images}. Our system estimates the figure-ground segmentation and detects the human body joints, semantically identifies the body parts, and reconstructs the 2d and 3d pose of the person. The design of recurrent multi-task loss functions at multiple stages of processing
supports the principled combination of the strengths of different 2d and 3d datasets,
without being limited by their different weaknesses. In experiments we perform ablation studies, evaluate the effect of various types of training data in the multitask loss, and demonstrate that state-of-the-art-results can be achieved at all processing levels. We show that, even in the wild, our monocular RGB architecture is perceptually competitive to state-of-the art commercial RGB-D systems.

\section{Acknowledgment}

This work was supported in part by CNCS-UEFISCDI under PCE-2011-3-0438, JRP-RO-FR-2014-16.



{\small
\bibliographystyle{ieee}

\begin{thebibliography}{10}\itemsep=-1pt

\bibitem{agarwal2006recovering}
A.~Agarwal and B.~Triggs.
\newblock Recovering 3d human pose from monocular images.
\newblock {\em PAMI}, 28(1):44--58, 2006.

\bibitem{akhter2015pose}
I.~Akhter and M.~J. Black.
\newblock Pose-conditioned joint angle limits for 3d human pose reconstruction.
\newblock In {\em CVPR}, 2015.

\bibitem{andriluka14cvpr}
M.~Andriluka, L.~Pishchulin, P.~Gehler, and B.~Schiele.
\newblock 2d human pose estimation: New benchmark and state of the art
  analysis.
\newblock In {\em CVPR}, 2014.

\bibitem{bo2010twin}
L.~Bo and C.~Sminchisescu.
\newblock Twin gaussian processes for structured prediction.
\newblock {\em IJCV}, 87(1-2):28--52, 2010.

\bibitem{bo2008fast}
L.~Bo, C.~Sminchisescu, A.~Kanaujia, and D.~Metaxas.
\newblock Fast algorithms for large scale conditional 3d prediction.
\newblock In {\em CVPR}, 2008.

\bibitem{bogo2016keep}
F.~Bogo, A.~Kanazawa, C.~Lassner, P.~Gehler, J.~Romero, and M.~J. Black.
\newblock Keep it smpl: Automatic estimation of 3d human pose and shape from a
  single image.
\newblock In {\em ECCV}, 2016.

\bibitem{bourdev10}
L.~Bourdev, S.~Maji, T.~Brox, and J.~Malik.
\newblock Detecting people using mutually consistent poselet activations.
\newblock In {\em ECCV}, 2010.

\bibitem{bulat2016}
A.~Bulat and G.~Tzimiropoulos.
\newblock Human pose estimation via convolutional part heatmap regression.
\newblock In {\em ECCV}, 2016.

\bibitem{carreira2015human}
J.~Carreira, P.~Agrawal, K.~Fragkiadaki, and J.~Malik.
\newblock Human pose estimation with iterative error feedback.
\newblock In {\em CVPR}, 2016.

\bibitem{chen2014}
X.~Chen and A.~L. Yuille.
\newblock Articulated pose estimation by a graphical model with image dependent
  pairwise relations.
\newblock In {\em NIPS}, 2014.

\bibitem{felzenszwalb2010object}
P.~F. Felzenszwalb, R.~B. Girshick, D.~McAllester, and D.~Ramanan.
\newblock Object detection with discriminatively trained part-based models.
\newblock {\em PAMI}, 32(9):1627--1645, 2010.

\bibitem{HernandezVelaS16}
A.~Hern{\'{a}}ndez{-}Vela, S.~Sclaroff, and S.~Escalera.
\newblock Poselet-based contextual rescoring for human pose estimation via
  pictorial structures.
\newblock {\em IJCV}, 118:49--64, 2016.

\bibitem{insafutdinov2016}
E.~Insafutdinov, L.~Pishchulin, B.~Andres, M.~Andriluka, and B.~Schiele.
\newblock Deepercut: A deeper, stronger, and faster multi-person pose
  estimation model.
\newblock In {\em ECCV}, 2016.

\bibitem{sminchisescu_locnpred09}
C.~Ionescu, L.~Bo, and C.~Sminchisescu.
\newblock Structural svm for visual localization and continuous state
  estimation.
\newblock In {\em ICCV}, 2009.

\bibitem{Ionescu14}
C.~Ionescu, J.~Carreira, and C.~Sminchisescu.
\newblock Iterated second-order label sensitive pooling for 3d human pose
  estimation.
\newblock In {\em CVPR}, 2014.

\bibitem{Ionescu14pami}
C.~Ionescu, D.~Papava, V.~Olaru, and C.~Sminchisescu.
\newblock {Human3.6M: Large Scale Datasets and Predictive Methods for 3D Human
  Sensing in Natural Environments}.
\newblock {\em PAMI}, 2014.

\bibitem{jia2014caffe}
Y.~Jia, E.~Shelhamer, J.~Donahue, S.~Karayev, J.~Long, R.~Girshick,
  S.~Guadarrama, and T.~Darrell.
\newblock Caffe: Convolutional architecture for fast feature embedding.
\newblock {\em arXiv preprint arXiv:1408.5093}, 2014.

\bibitem{Johnson10}
S.~Johnson and M.~Everingham.
\newblock Clustered pose and nonlinear appearance models for human pose
  estimation.
\newblock In {\em BMVC}, 2010.

\bibitem{johnson2010}
S.~Johnson and M.~Everingham.
\newblock Clustered pose and nonlinear appearance models for human pose
  estimation.
\newblock In {\em BMVC}, 2010.

\bibitem{kiefel2014human}
M.~Kiefel and P.~V. Gehler.
\newblock Human pose estimation with fields of parts.
\newblock In {\em ECCV}, 2014.

\bibitem{li20143d}
S.~Li and A.~B. Chan.
\newblock 3d human pose estimation from monocular images with deep
  convolutional neural network.
\newblock In {\em ACCV}, 2014.

\bibitem{li2015maximum}
S.~Li, W.~Zhang, and A.~B. Chan.
\newblock Maximum-margin structured learning with deep networks for 3d human
  pose estimation.
\newblock In {\em ICCV}, 2015.

\bibitem{liang2015human}
X.~Liang, C.~Xu, X.~Shen, J.~Yang, S.~Liu, J.~Tang, L.~Lin, and S.~Yan.
\newblock Human parsing with contextualized convolutional neural network.
\newblock In {\em ICCV}, 2015.

\bibitem{Lifshitz2016}
I.~Lifshitz, E.~Fetaya, and S.~Ullman.
\newblock Human pose estimation using deep consensus voting.
\newblock In {\em ECCV}, 2016.

\bibitem{Long15}
J.~Long, E.~Shelhamer, and T.~Darrell.
\newblock Fully convolutional networks for semantic segmentation.
\newblock In {\em CVPR}, 2015.

\bibitem{mori2006recovering}
G.~Mori and J.~Malik.
\newblock Recovering 3d human body configurations using shape contexts.
\newblock {\em PAMI}, 28(7):1052--1062, 2006.

\bibitem{newell2016}
A.~Newell, K.~Yang, and J.~Deng.
\newblock Stacked hourglass networks for human pose estimation.
\newblock In {\em ECCV}, 2016.

\bibitem{pfister2015flowing}
T.~Pfister, J.~Charles, and A.~Zisserman.
\newblock Flowing convnets for human pose estimation in videos.
\newblock In {\em ICCV}, 2015.

\bibitem{pishchulin2013}
L.~Pishchulin, M.~Andriluka, P.~Gehler, and B.~Schiele.
\newblock Poselet conditioned pictorial structures.
\newblock In {\em CVPR}, 2013.

\bibitem{pons2014}
G.~Pons{-}Moll, D.~J. Fleet, and B.~Rosenhahn.
\newblock Posebits for monocular human pose estimation.
\newblock In {\em CVPR}, 2014.

\bibitem{ramakrishna2012reconstructing}
V.~Ramakrishna, T.~Kanade, and Y.~Sheikh.
\newblock Reconstructing 3d human pose from 2d image landmarks.
\newblock In {\em ECCV}, 2012.

\bibitem{ramakrishna2014}
V.~Ramakrishna, D.~Munoz, M.~Hebert, J.~A. Bagnell, and Y.~Sheikh.
\newblock Pose machines: Articulated pose estimation via inference machines.
\newblock In {\em ECCV}, 2014.

\bibitem{rhodin2016general}
H.~Rhodin, N.~Robertini, D.~Casas, C.~Richardt, H.-P. Seidel, and C.~Theobalt.
\newblock General automatic human shape and motion capture using volumetric
  contour cues.
\newblock In {\em ECCV}, 2016.

\bibitem{rosales2001learning}
R.~Rosales and S.~Sclaroff.
\newblock Learning body pose via specialized maps.
\newblock In {\em NIPS}, 2001.

\bibitem{rosenhahn08}
B.~Rosenhahn, R.~Klette, and D.~Metaxas, editors.
\newblock {\em Human Motion, Understanding, Modelling, Capture and Animation},
  volume~36.
\newblock Springer Verlag, 2008.

\bibitem{ShakhnarovichVD03}
G.~Shakhnarovich, P.~A. Viola, and T.~Darrell.
\newblock Fast pose estimation with parameter-sensitive hashing.
\newblock In {\em ICCV}, 2003.

\bibitem{sigal07}
L.~Sigal, A.~Balan, and M.~J. Black.
\newblock Combined discriminative and generative articulated pose and non-rigid
  shape estimation.
\newblock In {\em NIPS}, 2007.

\bibitem{sigal2010}
L.~Sigal, A.~O. Balan, and M.~J. Black.
\newblock Humaneva: Synchronized video and motion capture dataset and baseline
  algorithm for evaluation of articulated human motion.
\newblock {\em IJCV}, 87(1-2):4--27, 2010.

\bibitem{simo2013joint}
E.~Simo-Serra, A.~Quattoni, C.~Torras, and F.~Moreno-Noguer.
\newblock A joint model for 2d and 3d pose estimation from a single image.
\newblock In {\em CVPR}, 2013.

\bibitem{sminchisescu_cvpr03}
C.~Sminchisescu and B.~Triggs.
\newblock Kinematic jump processes for monocular 3d human tracking.
\newblock In {\em CVPR}, 2003.

\bibitem{tang2016multi}
S.~Tang, B.~Andres, M.~Andriluka, and B.~Schiele.
\newblock Multi-person tracking by multicut and deep matching.
\newblock In {\em ECCV}, 2016.

\bibitem{Tekin2016}
B.~Tekin, A.~Rozantsev, V.~Lepetit, and P.~Fua.
\newblock Direct prediction of 3d body poses from motion compensated sequences.
\newblock {\em arXiv preprint arXiv:1402.0119}, 2015.

\bibitem{tompson2014joint}
J.~J. Tompson, A.~Jain, Y.~LeCun, and C.~Bregler.
\newblock Joint training of a convolutional network and a graphical model for
  human pose estimation.
\newblock In {\em NIPS}, 2014.

\bibitem{toshev2014}
A.~Toshev and C.~Szegedy.
\newblock Deeppose: Human pose estimation via deep neural networks.
\newblock In {\em CVPR}, 2014.

\bibitem{Wang_2014_CVPR}
C.~Wang, Y.~Wang, Z.~Lin, A.~L. Yuille, and W.~Gao.
\newblock Robust estimation of 3d human poses from a single image.
\newblock In {\em CVPR}, 2014.

\bibitem{wei2016}
S.~Wei, V.~Ramakrishna, T.~Kanade, and Y.~Sheikh.
\newblock Convolutional pose machines.
\newblock In {\em CVPR}, June 2016.

\bibitem{yang2013}
Y.~Yang and D.~Ramanan.
\newblock Articulated human detection with flexible mixtures of parts.
\newblock {\em PAMI}, 35(12):2878--2890, 2013.

\bibitem{Yasin2016}
H.~Yasin, U.~Iqbal, B.~Kruger, A.~Weber, and J.~Gall.
\newblock A dual-source approach for 3d pose estimation from a single image.
\newblock In {\em CVPR}, June 2016.

\bibitem{zhou2015sparseness}
X.~Zhou, M.~Zhu, S.~Leonardos, K.~Derpanis, and K.~Daniilidis.
\newblock Sparseness meets deepness: 3d human pose estimation from monocular
  video.
\newblock In {\em CVPR}, 2016.

\end{thebibliography}

}
\end{document}